**Shaping the Future through Innovations: From Medical Imaging to Precision Medicine**

Dorin Comaniciu[1], Klaus Engel[2], Bogdan Georgescu[1], Tommaso Mansi[1]

[1]Medical Imaging Technologies, Siemens Healthcare Technology Center, Princeton, New Jersey, USA
[2]Medical Imaging Technologies, Siemens Healthcare Technology Center, Erlangen, Germany

**Abstract:** *Medical images constitute a source of information essential for disease diagnosis, treatment and follow-up. In addition, due to its patient-specific nature, imaging information represents a critical component required for advancing precision medicine into clinical practice. This manuscript describes recently developed technologies for better handling of image information: photorealistic visualization of medical images with Cinematic Rendering, artificial agents for in-depth image understanding, support for minimally invasive procedures, and patient-specific computational models with enhanced predictive power. Throughout the manuscript we will analyze the capabilities of such technologies and extrapolate on their potential impact to advance the quality of medical care, while reducing its cost.*

## Introduction

Medical imaging has impacted the practice of medicine during the recent decades, contributing to greatly improved disease diagnosis, treatment and follow-up. Image-guided, minimally invasive procedures are becoming more and more common in hospitals, replacing conventional surgery and allowing faster recoveries with fewer post-procedure complications. We anticipate that this trend will continue, medical imaging playing an increasingly important role towards moving precision medicine into clinical practice. By being able to characterize anatomy, physiology and metabolism of the patient, medical imaging enables precise, personalized procedures and predictive, patient-specific therapy selection and delivery.

In this paper we highlight a number of technologies that we will most likely contribute to the success of medical imaging for the years to come, helping medical care to advance, while reducing its cost. In Section 1 we discuss Cinematic Rendering, a 3D visualization technology that is capable of producing superb photorealistic images from traditional Computer Tomography (CT) or Magnetic Resonance (MR) volumes, thus potentially enhancing the conspicuity of pathologies. Section 2 addresses the topic of next generation image understanding, which contributes to faster and more reproducible image reading, benefiting from the recent advances in machine learning and artificial intelligence. Furthermore, in Section 3 we discuss the real-time imaging needs in the Operating Room and focus on heart valve procedures, addressing both their planning and guidance. Finally, in Section 4 we present patient-specific computational models that contribute to advances in diagnosis, patient stratification, therapy selection and therapy optimization. All images shown in the paper are images of real, living patients.

1. **Cinematic Rendering: Photorealistic Visualization of Medical Images**

Efficient clinical decisions and procedures require the rapid appreciation of the relevant information contained within medical images. Even though medical image viewing based on multi-planar reconstruction (MPR) is still dominant in diagnostic imaging, the significance of three-dimensional visualization of medical data is rising. This is due to the fact that these methods allow much faster understanding of spatial anatomical structures and have the potential to increase the sensitivity and specificity of medical images. Especially medical professionals who are not trained in planar image viewing as well as patients benefit from such visualizations.

Recent advances in computer graphics have made interactive physically-based volume visualization techniques possible. Such techniques reproduce complex illumination effects in computer-generated images by mimicking the real-world interaction of light with matter. The results are physically plausible images that are often easier for the human brain to interpret, since the brain is trained to interpret the slightest shading cues to reconstruct shape and depth information. Such shading cues are often missing from computer generated images based on more simple geometric calculations such as ray casting.

We developed a physically-based volume rendering method called Cinematic Rendering [Engel 2016; Paladini 2015] which computes in real-time the interaction of visible photons with the scanned patient anatomy. The algorithm uses a Monte Carlo path tracing method to generate photorealistic or even hyper-realistic images by light transport simulation along hundreds or thousands of photons paths per pixel through the anatomy using a stochastic process (**Figure 1**)

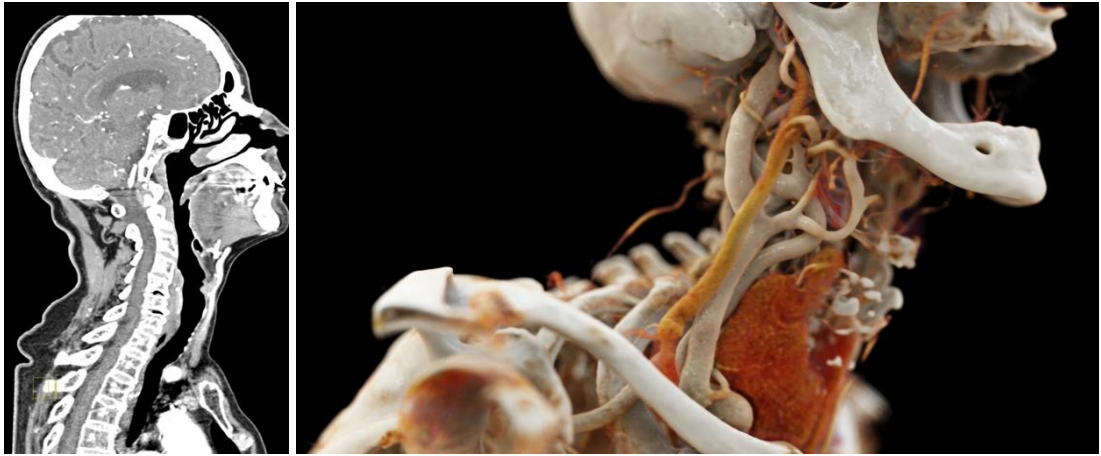

**Figure 1: Cinematic Rendering. Left: Original computed tomography (CT) data; Right: Cinematic rendering of the same dataset. Data courtesy of Israelitisches Krankenhaus, Hamburg, Germany.**

In traditional volume ray casting, only emission and absorption of radiant energy along a straight ray is considered. Radiant energy $q_e$ is emitted at each point $x'$ along the ray up to a maximum distance *D*.

$$L(x, \boldsymbol{\omega}) = \int_0^D e^{-\tau(x,x')} q_e(x') dx' \qquad (1)$$

The emitted radiant energy at each point is absorbed according to the Beer-Lambert law along the ray to the observer location with absorption coefficients $\sigma_a$.

$$\tau(x, x') = \int_x^{x'} \sigma_a(t) dt \qquad (2)$$

Single scattering is usually modelled in traditional volume rendering using a surface shading model that considers local gradient information of the volume data (local illumination). While this integral can be easily solved numerically using a Riemann integral, the method neglects complex light paths with multiple scattering events and extinction of light (global illumination).

In contrast, the Monte Carlo path tracing integration method solves the following multi-dimensional and non-continuous rendering equation:

$$L(x,\boldsymbol{\omega}) = \int_0^D e^{-\tau(x,x')}\sigma_S(x')\left[\int_{\Omega_{4\pi}} p(\boldsymbol{\omega},\boldsymbol{\omega}')L_i(x',\boldsymbol{\omega}')d\boldsymbol{\omega}'\right]dx' \quad (3)$$

Equation (3) determines the radiant flux (radiance) *L* at distance x received from the direction $\boldsymbol{\omega}$ along a ray. We have to integrate the radiance scattered into that direction from all possible directions $\boldsymbol{\omega}'$ at all points along the ray up to a maximum distance *D*. The optical properties of a relevant tissue are defined using the phase function $p(\boldsymbol{\omega},\boldsymbol{\omega}')$, which describes the fraction of light travelling along a direction $\boldsymbol{\omega}'$ being scattering into the direction $\boldsymbol{\omega}$. $L_i(x',\boldsymbol{\omega}')$ is the radiance arriving a distance $x'$ from direction $\boldsymbol{\omega}'$. In practice, we model scattering in different tissue types using a Henyey-Greenstein phase function and compute shading of implicit surfaces using a BRDF (bidirectional reflectance distribution function). Radiance scattering into the direction $\boldsymbol{\omega}$ is also absorbed and scattered out of the direction $\boldsymbol{\omega}$. This is modelled using the optical depth $\tau$, with extinction coefficient $\sigma_t = \sigma_s + \sigma_a$, defined as the sum of scattering ($\sigma_s$) and absorption ($\sigma_a$) coefficients:

$$\tau(x,x') = \int_x^{x'} \sigma_t(t)dt \quad (4)$$

Note that, in contrast to out-scattering, absorption and in-scattering, emission was omitted in the rendering equation for simplicity. Since the rendering equation cannot be computed analytically, solving the integral numerically would involve sampling the function at many distances, each with many directions. Additionally, $L_i$ must be computed with the same rendering equation to allow multiple scatter events. Since this would be computationally too complex, the Monte Carlo method allows us to compute the radiance at random positions along the ray with light being in-scattered from random directions. By averaging many of such Monte Carlo samples into a single image we can progressively generate a smooth final result. By means of multiple sampling, the convergence of the method can be accelerated considerably.

The medical data is illuminated using image-based lighting by high-dynamic range lighting environments, which can either be captured photographically or generated synthetically. Photographically captured lighting leads to a very natural appearance of the data when compared to images created using the traditional ray casting method. Such natural lighting in combination with the accurate simulation of photon scattering and absorption, leads to photorealistic images (see **Figure 1**) that resemble many shading effects that can be observed in nature, such as soft shadows, ambient occlusion, volumetric scattering and subsurface photon interaction. By modelling a virtual camera with variable aperture, focal length and exposure, additional effects such as depth-of-field and motion blur can be produced. Motion blur allows movies generated using our key frame animation engine to be smoother during fast camera movements while, similar to photography, depth-of-field effects allow to focus the attention of a viewer on a particular structures.

Beyond photorealism the algorithm also permits to visualize invisible or hidden processes such as the propagation of electrical activation on the heart surface or metabolic processes in the body. Such hyper-realistic images are created by modelling visible light photon emission from voxels affected by electrical activation, increased metabolism indicated by Positron Emission Tomography (PET) or the detection of chemical compounds such as monosodium urate from a dual-energy CT scan (**Figure 2**).

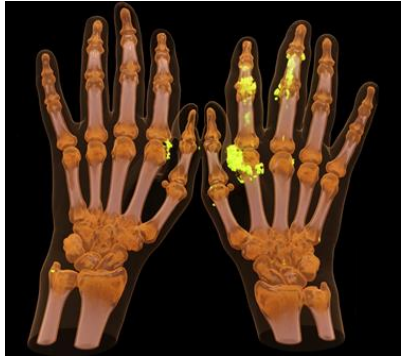

**Figure 2: Gout visualization. Modelling photon emission from urate detection by a dual-energy CT scan.**

The combination of different imaging modalities in a single picture, such as PET, MR and CT as well as simulated and computed data provides important flexibility to show the spatial relation of anatomical structures and functional data (see **Figure 3**).

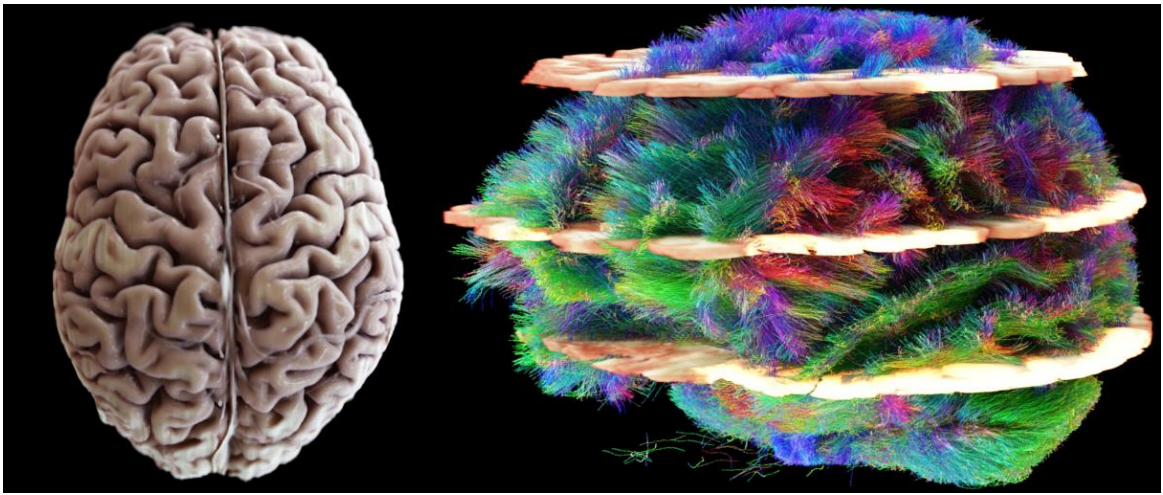

**Figure 3: Human brain visualization. Left: Cinematic Rendering of a Magnetic Resonance (MR) image of the brain acquired with a 7T scanner. Data courtesy of Max Planck Institute, Leipzig, Germany. Right: Cinematic Rendering of three anatomical MR slices, functional MR Imaging (fMRI), and fiber data computed from Diffusion Tensor Imaging (DTI) of the brain. The activation of the speech center captured by fMRI is modelled using light emission and results in the yellow glowing lighting effect on the anatomy.**

Another important application of Cinematic Rendering is the visualization of dynamic processes from 4D CT or MR scans in combination with time-dependent data from simulations. All such data sources can be combined frame-by-frame and played using an animation engine to create photorealistic movies which allow conveying an effective clinical message to the target audience.

While diagnostics will certainly still rely on traditional planar reconstruction based visualization methods, we have strong indications that special diagnostic applications might benefit from the flexibility and expressiveness of the new Cinematic Rendering technology. For instance, a robust demand for such visualization methods can be seen for surgery planning and intraoperative imaging, where a good spatial understanding of the anatomy and processes in the human body is required (**Figure 4**).

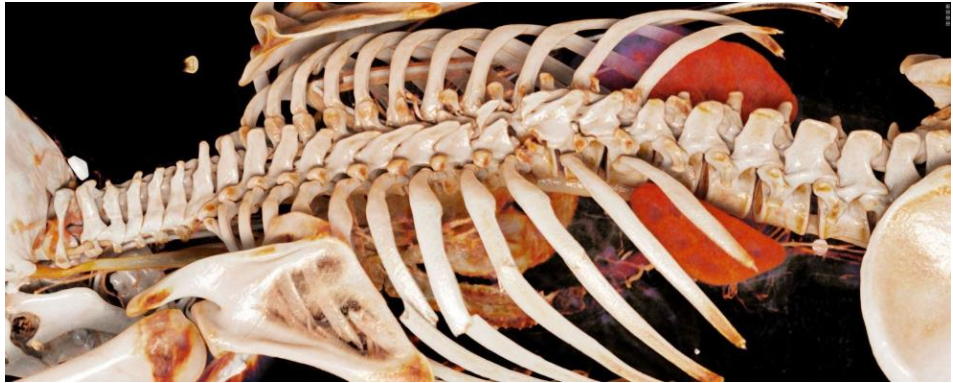

**Figure 4: Polytrauma visualization. Cinematic Rendering of a polytrauma patient with multiple spinal and costal fractures. CT image data courtesy of Vancouver General Hospital, Canada.**

Furthermore, the use of hyper-realistic imaging for anatomical education of medical students as well as the general public is obvious. And finally, such images are ideal for the efficient communication of findings, diagnoses and surgery results, either among medical professionals or to patients, potentially increasing trust in clinical decisions and procedures.

## 2. Artificial Intelligence and Image Understanding

Handling the complexity of medical images involves understanding thousands of anatomical classes and concepts, while numerous relationships are necessary to symbolically represent the phenotypic structure of the human body [FMA, 2012]. There are multiple dimensions along which this information can be structured, for example one can look at the human anatomy from a regional point of view (limbs, head), or constitutional point of view (lymphatic duct, skin) or system (nervous, cardiovascular, musculoskeletal). Such ontology-based or symbolic representation is a form that is understandable by humans and it is also navigable, parseable and interpretable by machine-based systems.

Fast and robust anatomical concept extraction is a fundamental task in medical image analysis that supports the entire workflow from diagnosis, patient stratification, therapy planning, intervention and follow-up. Current state-of-the art solutions are based on machine learning, being enabled by the availability of large annotated medical databases and the increased computational capabilities [Zheng and Comaniciu, 2014]. Typical methods use example images of the anatomy of interest to learn a classifier that will be able to discriminate between inputs that contain the target anatomy or something else. Such classifiers can be used to automatically label images, detect landmarks or segment the target object (see **Figure 5**).

For example, in the context of object detection, the classifier is scanned over all possible values of the parameter space (say translation $T$, rotation $R$ and scale $S$) to find the high probability regions that will correspond to object location. This is done by using a classifier that will approximate the probability $p(T, R, S|I)$ for an image $I$, where the classifier is trained with object image features for one class and non-object image features for the other class. At runtime the object is determined by regions of the parameter space $<\hat{T}, \hat{R}, \hat{S}>$ with high probability: $\operatorname{argmax}_{T,R,S} p(T, R, S|I) = <\hat{T}, \hat{R}, \hat{S}>$.

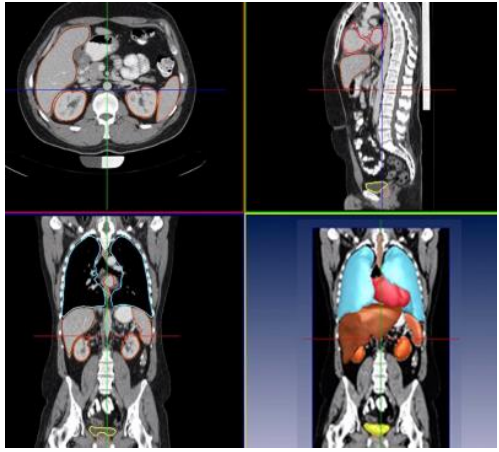

**Figure 5: Identifying and segmenting anatomical structures in a whole body CT scan (only partial views are shown). Tens of anatomical landmarks are being identified and multiple organs such as heart, lungs, liver, kidneys, spleen, prostate and bladder are segmented and quantified [Seifert 2009]**

In the past years, we have developed technologies such as Marginal Space Learning (MSL) [Zheng and Comaniciu, 2014] for automated and efficient scanning of medical images. The MSL method consists in learning image-based classifiers in high probability marginal spaces of the object parameterization. A set of classifiers are trained in stages: first in the translation hypotheses space $\Omega_T(I)$, where $I$ is the image, next the augmented translation-orientation hypotheses space $\Omega_{T,R}(I)$ and finally in the translation-orientation-scale hypotheses space $\Omega_{T,R,S}(I)$. The spaces are constructed by augmenting the high probability hypotheses with all the possible discrete values of the next parameter space such as:

$$\operatorname{argmax}_T p(\Omega_T|I) = \Omega_{\hat{T}} \xrightarrow{\text{Augment with discrete rotations}} \Omega_{\hat{T},R} \, .$$

$$\operatorname{argmax}_{T,R} p(\Omega_{\hat{T},R}|I) = \Omega_{\hat{T},\hat{R}} \xrightarrow{\text{Augment with discrete scales}} \Omega_{\hat{T},\hat{R},S} \, . \tag{5}$$

$$\operatorname{argmax}_{T,R,S} p(\Omega_{\hat{T},\hat{R},S}|I) = \Omega_{\hat{T},\hat{R},\hat{S}}.$$

The MSL technique is generic and can be extended to any type of parameterized spaces. We used MSL to automatically determine hundreds of landmarks, segment, track and quantify all main organs, delineate and index the vascular tree, brain structures and the skeleton.

Recent advances in machine learning and artificial intelligence have created end-to-end learning architectures where all stages of the processing are jointly optimized. For example, representation learning with Deep Neural Networks (DNN) enables automatic extraction of representative image features without the need of feature engineering [Zheng 2015]. DNN allow learning complex patterns from very large heterogeneous image databases. We have recently introduced Marginal Space Deep Learning (MSDL) [Ghesu 2016a] that combines the strength of automated feature design of DNN with efficient learning in marginal spaces. In MSDL (**Figure 6**) the classifier is trained directly on parameterized image patches and used to estimate the probability distribution: $\mathcal{R}(\Omega(I); w, b) \approx p(\Omega(I)|I)$ where $\mathcal{R}$ is a deep neural network response function parameterized by the weights $w$ and biases $b$ of each layer. In addition, more efficient scanning of DL networks is achieved through network approximation (sparsification) techniques by minimizing the residual $\|\mathcal{R}(\Omega(I); w, b) - \mathcal{R}(\Omega(I); w_s, b_s)\|$, where $w_s$ and $b_s$ are the weight and bias of the approximated sparse network $\mathcal{R}(\Omega(I); w_s, b_s)$. As a result, $\mathcal{R}(\Omega(I); w_s, b_s)$ has much fewer parameters and/or access much less data from the image making possible scanning for parameterized objects in 3D or 4D images. With MSDL, we have shown significant performance improvements in terms of both accuracy and speed on aortic valve detection in volumetric ultrasound and landmark detection in CT scans [Ghesu 2016a].

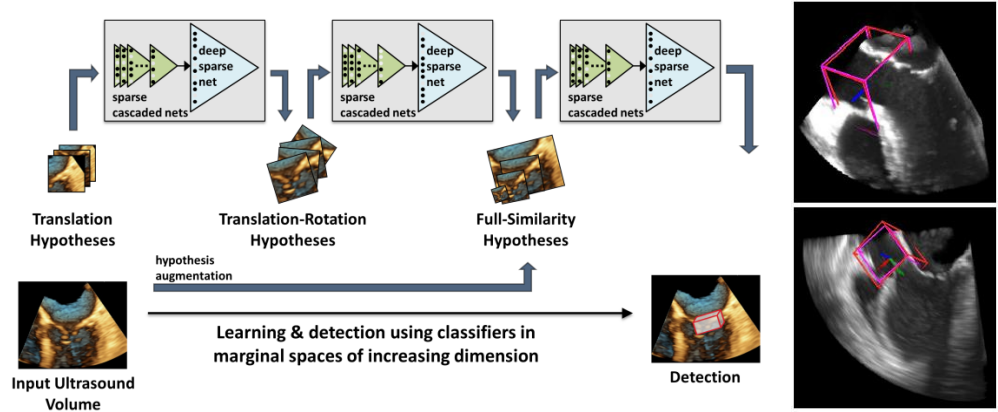

**Figure 6: Marginal Space Deep Learning. Left: Learning in increasingly dimensional spaces focused on high probability regions with deep neural sparse networks. Right: Example of aortic-valve detection in volumetric ultrasound.**

With the goal to add more intelligence into image analysis, our recent work has focused on artificial intelligence agents that can be trained using Deep Reinforcement Learning (DRL) techniques to simultaneously model both the object appearance and the object search strategy as a unified behavior [Ghesu 2016b]. The idea is to train an agent that can navigate within an image to find an anatomy of interest. In other words, the agent learns automatically optimal paths that converge to the target object, thus eliminating the need for exhaustive search [Ghesu 2016b].

A typical Reinforcement Learning (RL) technique is modeled as a Markov Decision Process (MDP) defined on the tuple ($S$, $\mathcal{A}$, $Tr$, $r$, $\gamma$), where $S$ represents a finite set of agent states, $\mathcal{A}$ represents a finite set of actions that the agent can perform to interact with the environment, $Tr: S \times \mathcal{A} \times S \to [0,1]$ is a stochastic transition function between two states by performing a specific action, $r: S \times \mathcal{A} \times S \to \mathbb{R}$ is the scalar reward expected after a state transition and $\gamma$ is a future rewards discount factor. One target in RL is to find an optimal of the action-value function $Q^*(s,a): S \times \mathcal{A} \to \mathbb{R}$ that corresponds to the maximum expected future rewards when performing action $a$ in state $s$: $Q^*(s,a) = \max_\pi E[r_t|s_t = s, a_t = a, \pi]$ where $\pi$ represent an action policy that determines the behavior of the agent. For object detection, such an agent can be trained by having $S$ as the current estimates of the object parameters given the image (e.g. spatial coordinates), $\mathcal{A}$ the discrete steps of parameter changes and a reward system that is related to how close the agent gets to the target by performing the actions. Given the model definition, a DNN can be trained to approximate the optimal action-value function $Q^*$ directly from the image values parameterized by the current state or object parameters. The optimal action-value function implicitly defines the optimal policy $\pi^*$ which guides the agent in finding the target object. This paradigm, where the agent simultaneously learns an object model and how to use the model, can be extended to a wide variety of image parsing actions (**Figure 7**).

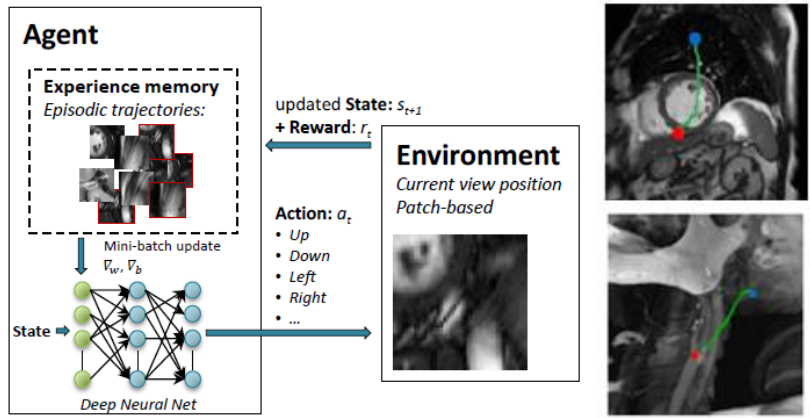

**Figure 7: Artificial agent for landmark detection. Left: Interaction of an artificial agent with the environment for detecting anatomical landmarks: Current state is defined by the image window and the agent performs the optimal action according to the learned behavior, which results in a new state and reward feedback. Right: 2D and 3D paths for detecting anatomical landmarks (in blue: starting point of the agent, red: target).**

Combining these types of learning techniques and classifiers with ontology based representation allows for semantic navigation of all the available patient data [Seifert 2011]. In particular, expanding end-to-end learning systems to include medical knowledge with powerful learning-based representation and reasoning systems will allow building hierarchical representations dynamically based on the current task. This will facilitate comprehensive automated analysis and reporting based on integrated imaging and non-imaging information, past reports and embedded medical knowledge. For instance, these methods will also enable integrated analysis of patient history with support for semantic search and case comparison by finding similar cases and treatments with related clinical knowledge and guidelines. Finally, the method facilitates knowledge sharing and population analytics. Such information empowers the radiologist towards increased efficiency and reduced uncertainty.

## 3. Support for Minimally-Invasive Procedures

Progress in medical imaging technologies and image analysis are making increasingly complex minimally invasive procedures possible. Techniques like heart valve repair or replacement can now be performed percutaneously, a relevant example being Transcatheter Aortic Valve Replacement (TAVR). Effective execution of minimally invasive procedures strongly relies on medical imaging.

First, the devices need to be selected to fit patient's anatomy. To that end, quantitative imaging is used to accurately measure the anatomy under consideration. For instance, the size of TAVR devices is determined from the dimensions of the aortic root. This is usually performed on CT data, but for patients suffering from kidney failure, novel, full-volume, real-time 3D TEE (trans-esophageal echocardiography) now enables the quantification of heart valves and blood flow without contrast agent. Fully exploiting this new imaging modality, we recently developed an advanced machine learning technology to estimate a personalized model of the heart valves (**Figure 8**). In brief, the algorithm, based on Marginal Space Learning, first detects the region of interest (ROI) where the valve is located. Within that ROI, key anatomical landmarks are detected and a parameterized triangulated surface is fitted to model the patient's valve. The detectors of each stage of the algorithm are trained from a large database of annotated images. The method is generic, and has been applied to other imaging modalities, like CT [Ionasec 2010].

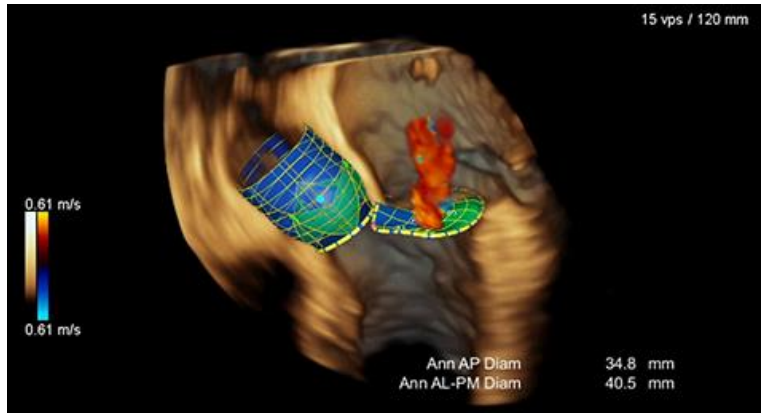

**Figure 8: Real-time 3D TEE and 3D Doppler. New TEE imaging allows full-volume, real-time visualization of cardiac anatomy (B-mode) and blood flow (color Doppler). For the first time, the clinician can visualize jointly the valve anatomy and potential insufficiency.**

Second, the operator needs image guidance to effectively deliver the device. Angiography is the modality of choice in the hybrid OR as it allows the visualization of catheters in real-time. However, soft tissues are hard to distinguish in these images. Advanced navigation concepts based on augmented reality have thus been investigated to enhance the angiography images with overlay of the targeted organ. 3D preoperative images are registered to the 2D scene (often facilitated by the injection of contrast) using pattern matching, multi-organ registration or multi-view reconstruction. Supported by imaging technologies, new minimally invasive procedures are emerging, like the recent transcatheter mitral valve replacement repairs techniques, which in turns require unprecedented levels of registration accuracy and robustness to cope with moving devices and image artifacts.

The future will likely go towards real-time 3D guidance, as hybrid solutions combining angiography and 3D TEE are becoming available. Through TEE probe pose estimation in the 3D space from angiography images, real-time ultrasound images can be automatically registered to the angiography space. User-defined landmarks, anatomical models or the ultrasound images directly can then be overlaid to the angiography image to guide the cardiologist towards the target (**Figure 9**). To reach the accuracy and speed requirements for real-time intervention guidance, we recently introduced a 3D TEE probe pose estimation based on deep learning [Miao 2016], yielding high accuracy at a frame-rate of 15fps.

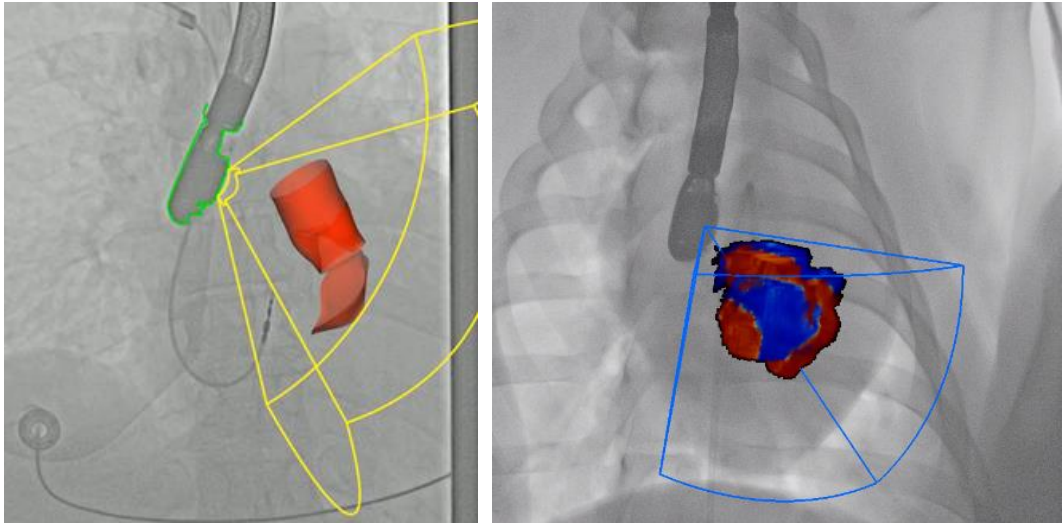

**Figure 9: 3D TEE – Angiography fusion. Left: Real-time overlay on angiography image of the valve model estimated from 3D TEE image acquired at the same time (see Figure 8); Right: Real-time overlay of the 3D Doppler generated by the 3D TEE probe.**

## 4. Decision Support through Patient Specific Computational Models

While increasingly robust and accurate quantification methods are being available, new solutions based on computational models of human physiology are being investigated to extract more physiological information from the images and facilitate patient-specific planning through predictive algorithms.

A first example is Fractional Flow Reserve (FFR), the current gold standard parameter that characterizes coronary stenosis severity. In standard of care, FFR is measured invasively using pressure catheters. During the past years, we have developed non-invasive, image-based FFR methods ($cFFR_{CFD}$) based on CT images and reduced-order computational fluid dynamics (CFD) models, making it possible to calculate FFR at the bed-side [Sharma 2012] with excellent performance. Furthermore, with the advances in machine learning, we have demonstrated that it is now possible to calculate FFR non-invasively in seconds on a standard workstation. Based on deep learning, the new approach consists in learning the CFD model directly from anatomical features [Itu 2016]. First, a database of 12,000 coronary geometries with more than 1,000,000 coronary segments was computed, with randomly positioned stenosis. Second, a reduced-order CFD model was used on all 12,000 geometries to calculate the resulting FFR. Both heart and systemic circulation models were included for proper boundary conditions. Third, a deep-network ($cFFR_{ML}$) was trained to predict the FFR value given geometric features computed upstream, at and downstream the stenosis. Tested on 127 unseen lesions from 87 patients, $cFFR_{ML}$ could be calculated in 2.4 seconds in average, with a correlation coefficient of 0.9994 ($p<0.001$) and no bias with respect to $cFFR_{CFD}$. Compared to the invasive FFR value, $cFFR_{ML}$ sensitivity was 81.6%, specificity 83.9% and accuracy 83.2%, achieving similar performance as other non-invasive FFR methods (**Figure 10**).

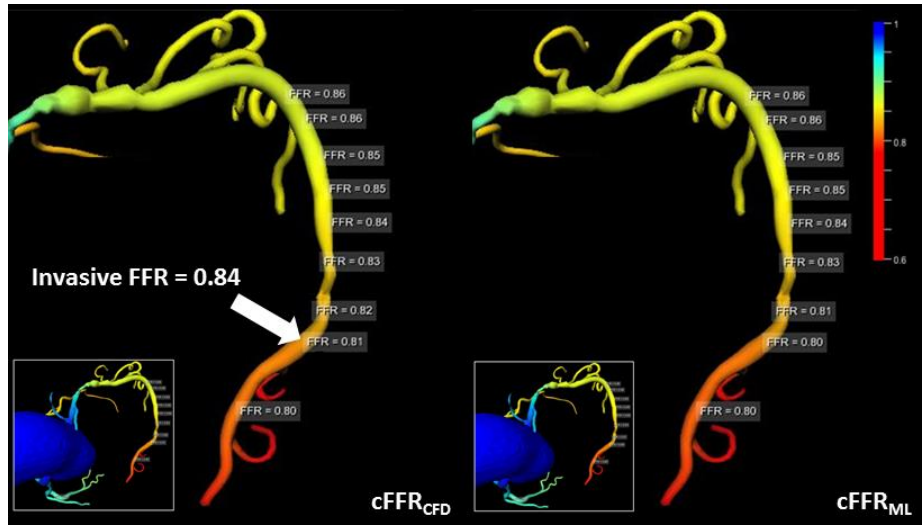

**Figure 10. Testing deep learning based FFR. Left: cFFR$_{CFD}$ computed using reduced order CFD model; Right: cFFR$_{ML}$ values. The deep learning approach could capture the FFR values accurately, while reducing the calculations to seconds.**

Similarly, we have developed machine learning techniques to estimate a patient-specific model of cardiac electrophysiology (EP) for cardiac resynchronization therapy (CRT) planning [Zettinig 2014, Kayvanpour 2015]. In particular, in [Neumann 2016], we introduce an intelligent agent trained following RL concepts to to estimate cardiac electrical conductivities directly from 12-lead ECG. Following the notations introduced in Section 2, we defined the RL Markov Decision Process ($S, A, Tr, r, \gamma$) as follows. The states $s \in S$ are discretized objective values (absolute difference between target and current QRS duration and electrical axis). The actions $a \in A$ are to increment and decrement the left endocardial, right endocardial, and myocardial conduction velocity (six actions in total). In a first stage, the agent learns through random exploratory simulations the state transition probabilities $Tr$, which encode how the model behaves when the electrical conductivities changes. Given this knowledge, the best personalization strategy is learned through RL. The reward $r$ is defined such that at every step $t$, starting from a state $s_t$, the agent receives a negative reward for all actions $a_t \in A$ except for the one $a_t^*$ that leads it to the parameters for which the EP model best matches the observed ECG. Finally, the parameter $\gamma$ was set to 0.9 or higher to favor long-term rewards (and thus global optimum). Tested on 83 consecutive patients, the artificial agent could achieve similar goodness of fit as a hand-crafted, state-of-the-art optimization method [Neumann 2016], while being 2.5 times faster.

Such personalized models can be used for patient-specific therapy planning. For instance, a user could use the model to virtually test different CRT pacing protocols [Kayvanpour 2015]. She would virtually place the CRT leads, program the virtual device, and update the model to visualize the impact of the CRT pacing on cardiac function (**Figure 11**), thus increasing the confidence on the therapy and the best strategy to apply to the patient under consideration.

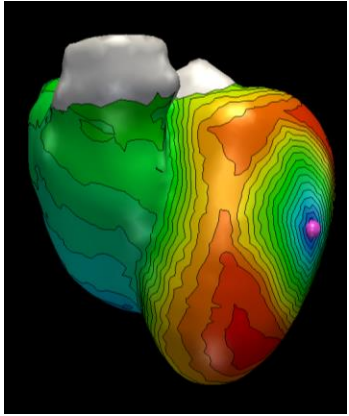

**Figure 11. Personalized CRT planning. Effect of ventricular pacing on cardiac electrophysiology as calculated by an individualized computational model of heart function**

## 5. Conclusion

In this paper we discussed recent technologies that will most likely make an important impact on medical imaging. Techniques like Cinematic Rendering will help increasing the sensitivity and specificity of images, by enhancing the pathology conspicuity. Advanced image understanding will streamline the image measurements and image interpretation, by increasing the speed of reading, while introducing more reproducibility in the system. The new heart valve technologies, based for instance on 3D TEE/Doppler, will help a more precise characterization of the patient's anatomy in the OR, while the 3D TEE – Angiography Fusion will support better guidance. Finally, patient specific computational modeling opens the door to a new generation of decision support systems that help clinical decision making not only by integrating and analyzing data from different sources, but also by modeling both the anatomy and function of the patient, thus exhibiting enhanced predictive power.

## Acknowledgement

The authors would like to thank all colleagues at the Siemens Healthcare Technology Center who contributed to the works discussed in this article. We would also like to thank our clinical and non-clinical collaborators for the strong support over many years.

## Disclaimer